\documentclass[conference]{IEEEtran}

\usepackage{multirow}
\usepackage{booktabs}

\usepackage{times}

\usepackage[numbers,sort&compress]{natbib}

\usepackage{multicol}
\usepackage[bookmarks=true]{hyperref}

\usepackage{pdfsync}
\usepackage[subrefformat=parens,labelformat=parens]{subfig}
\usepackage{graphicx}
  \graphicspath{{figure/}}
  \DeclareGraphicsExtensions{.pdf,.jpg,.png,.eps}
\usepackage{pdfpages}
\usepackage{bm}
\usepackage{amsmath,amssymb}
\usepackage{algorithm}
\usepackage{algorithmicx}
\usepackage{algpseudocode}
\usepackage{xspace}
\usepackage{booktabs}
\usepackage{soul}

\usepackage{todonotes}

\newcommand{\letsdrive}{LeTS-Drive\xspace}
\newcommand{\Pomdpbaseline}{\textit{Decoupled-Action-POMDP}\xspace}
\newcommand{\Hypbaseline}{\textit{Joint-Action-POMDP}\xspace}
\newcommand{\Ilbaseline}{\textit{\letsdrive-Imitation}\xspace}
\newcommand{\Porcabaseline}{\textit{NH-PORCA}\xspace}
\newcommand{\pomdpbaseline}{Decoupled-Action-POMDP\xspace}
\newcommand{\hypbaseline}{Joint-Action-POMDP\xspace}
\newcommand{\ilbaseline}{\letsdrive-Imitation\xspace}
\newcommand{\porcabaseline}{NH-PORCA\xspace}

\newcommand{\secref}[1]{Section~\ref{#1}}
\renewcommand{\eqref}[1]{(\ref{#1})}
\newcommand{\figref}[1]{Fig.~\ref{#1}}

\newcommand{\tabref}[1]{Table~\ref{#1}}

\newcommand{\ie}{\textrm{i.e.}}
\newcommand{\eg}{\textrm{e.g.}}

\DeclareMathOperator*{\argmax}{arg\,max} 

\newlength{\citeskipup}
\newlength{\citeskipdown}

\usepackage{color}
\definecolor{fullred}{rgb}{0.95,.0,.1} 
\newcounter{cmt}

\newcommand{\speed}{\ensuremath{v\xspace}} 

\newcommand{\nscen}{\ensuremath{K}\xspace}

\newcommand{\scenario}{\ensuremath{{\boldsymbol{\phi}}}\xspace}

\newcommand{\scenariosetnode}[1]{\ensuremath{{\boldsymbol{\Phi}_{#1}}}\xspace}

\newcommand{\randnum}{\ensuremath{\varphi}\xspace}

\newcommand{\act}{\ensuremath{a}\xspace}

\newcommand{\sinit}{\ensuremath{s_0}\xspace}

\newcommand{\node}{\ensuremath{b}\xspace}

\newcommand{\rootnode}{\ensuremath{b_0}\xspace}

\newcommand{\nvisit}[1]{\ensuremath{N({#1})}\xspace}
\newcommand{\nvisita}[2]{\ensuremath{N({#1},{#2})}\xspace}
\newcommand{\ubsymbol}{\ensuremath{u}\xspace}

\newcommand{\uba}[2]{\ensuremath{\ubsymbol({#1},{#2})}\xspace}

\newcommand{\lbsymbol}{\ensuremath{l}\xspace}

\newcommand{\accel}{{\small \textsc{Accelerate}}}
\newcommand{\decel}{{\small \textsc{Decelerate}}}
\newcommand{\maintain}{{\small \textsc{Maintain}}}

\newcommand\fw{0.16\textwidth}

\def \Fbox #1{\fbox{#1}}

\usepackage{hyperref}

\newcommand{\subfig}[1]{\textit{#1}}

\usepackage{fancyhdr}
\title{\LARGE \bf
\letsdrive: Driving in  a Crowd by Learning from Tree Search
}

\author{Panpan Cai, Yuanfu Luo, Aseem Saxena, David Hsu, Wee Sun Lee\\
School of Computing, National University of Singapore, 117417 Singapore}

\usepackage[left=0.625in, right=0.625in, top=0.75in, bottom=1in]{geometry}

\begin{document}

\maketitle
\thispagestyle{fancy}
\pagestyle{empty}

\begin{abstract}
  Autonomous driving in a crowded environment, \eg, a busy traffic
  intersection, is an unsolved challenge for robotics. The robot vehicle must
  contend with a dynamic and partially observable environment, noisy sensors, and many agents.  A
  principled approach is to formalize it as a Partially Observable Markov
  Decision Process (POMDP) and  solve it through online belief-tree search.  To
  handle a large crowd and achieve real-time performance in this very
  challenging setting, we propose \letsdrive, which integrates online POMDP
  planning and deep learning. 
  It  consists of two phases. In the offline phase, we learn a policy
  and the corresponding value function by imitating the belief tree search. In
  the online phase, the learned policy and value function guide the belief
  tree search.
  \letsdrive leverages the robustness of planning and the
  runtime efficiency of learning to enhance the performance of both.
  Experimental results in simulation  show that \letsdrive outperforms either planning or imitation learning alone and develops sophisticated driving
  skills.


\end{abstract}

\section{INTRODUCTION}

\begin{figure}
\centering
\includegraphics[width=\columnwidth]{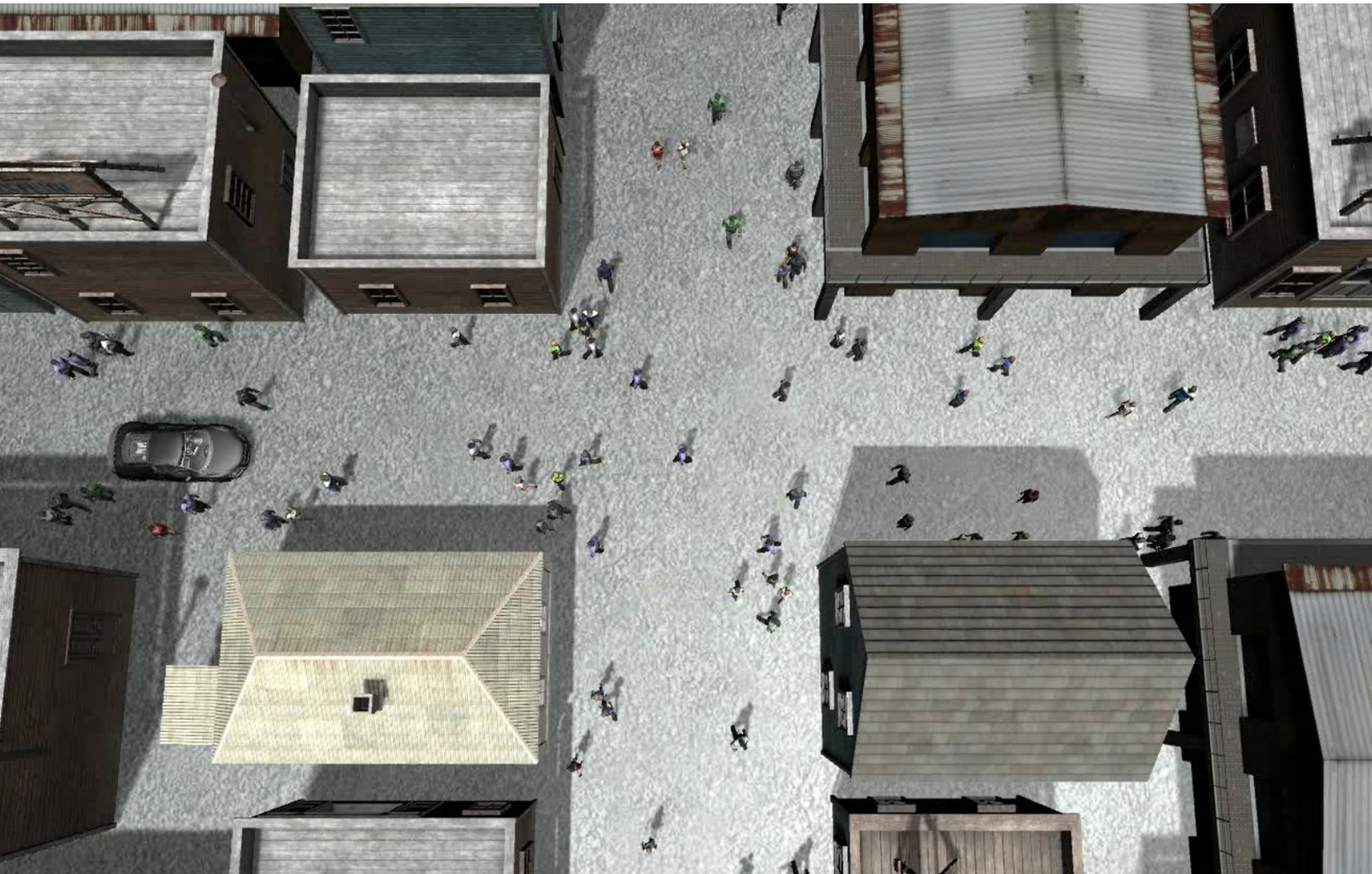}
\caption{
Autonomous driving in a crowd.  A robot vehicle drives amidst many moving
pedestrians. Each pedestrian moves towards  his/her own goal while interacting
with other pedestrians and the vehicle. 
}
\label{fig:Driving}
\end{figure}

Experienced humans can drive in crowded streets, markets, or squares without colliding with any others and make their way efficiently through the crowd. However, robot vehicles can easily fail in these scenarios. The environment is highly dynamic, comprising many agents interacting with each other, while the robot only has limited perception capabilities. 
Autonomous driving in a crowd remains an open problem. 
This paper studies a representative of such problems (\figref{fig:Driving}): driving amidst many moving pedestrians in a map-constrained environment. 
Typical approaches for crowd driving using local collision avoidance \cite{ORCA, PORCA} not only generate jerky and zig-zags motions, but can also be easily trapped in local optima, resulting in the vehicle getting stuck in the crowd. Successful driving requires more sophisticated skills, \eg, detouring to by-pass pedestrians or inching forward to make space in a dense crowd. The key here is to perform long-term planning: predict the motion of pedestrians for multiple steps and plan for the vehicle accordingly. 
However, the planning problem is extremely challenging. 
A planning algorithm needs to reason in a high-dimensional, partially observable state space formed by surrounding pedestrians, and needs to model a plethora of uncertainties: noisy sensing, unknown intentions of pedestrians, and complex interactions between them. Failing to handle these uncertainties leads to severe or even fatal accidents.

A principled approach for planning under uncertainty is the Partially Observable Markov Decision Process (POMDP). POMDP captures partially observability in a \textit{belief}, which is a probability distribution over states, and reasons about the stochastic effects of robot actions, sensor information, and environment dynamics on the belief. 
Complex problems require online planning: perform a look-ahead search in a \textit{belief tree} to compute a \textit{policy}, execute the first action in the policy, and re-plan at each time step.
However, the computational complexity of belief tree search grows polynomially with the size of the vehicle action space and exponentially with the number of surrounding pedestrians and the planning horizon.

To handle the complexity of driving a vehicle among many pedestrians, state-of-the-art planning methods utilize two approaches: restricting POMDP to control only the accelerations along a pre-planned path \cite{Bai_2015}, and massively parallelizing the planning with GPUs \cite{Hyp-despot}. These approaches achieved impressive performance in driving through real crowds of pedestrians. However, we argue that only controlling accelerations is too restrictive for long-term strategies required in crowded streets of cities like Beijing, Bombay, or Hanoi. We seek to relax this constraint.

In this paper, we scale up POMDP planning to much larger search spaces by integrating it with learning. The idea is to use a learned policy to represent prior knowledge and use planning to further optimize the policy for a particular problem instance.
Our method learns two deep neural networks to help the search: a policy network that guides search to useful parts of the search space, and a value network to estimate the value of belief subtrees without having to search them. We further exploit the domain knowledge in planning to design our deep neural networks: we use a Gated Path Planning Network (GPPN) \cite{GPPN}, a neural network that approximates the value iteration algorithm, to provide an initial plan, which is refined by another neural network component.

Our proposed method, \letsdrive, first executes imitation learning \cite{hussein2017imitation} using an existing belief tree search algorithm \cite{Hyp-despot} as an expert to generate training data. It then uses the learned policy and value functions as heuristics to guide the belief tree search, which generates actions to drive the vehicle.
\letsdrive exploits the robustness of planning and the runtime efficiency of learning, and integrates them to advance the capabilities of both. 
On one hand, \letsdrive uses the learned policy to characterize the robot's long-term behaviors and avoid searching a deep tree. On the other hand, it exploits a prediction model at hand to optimize the robot's short-term behaviors, and in the meantime, corrects possible mistakes raised by neural networks. 

\letsdrive is inspired by the AlphaGO \cite{AlphaGO} that integrates Monte
Carlo tree search (MCTS) with neural networks to learn board games like GO, Shogi,
and chess from self-play. However, crowd driving is substantially different and much more challenging than board games. First, instead of competing with one opponent, the robot vehicle interacts with many agents who further interact with each other. In this case, it is intractable to search over the actions of all agents. 
Instead, we predict pedestrians' motion by reasoning about their intended navigation goals and modeling their reciprocal interactions.
Second, states in board games are fully observable and action executions are deterministic. However, in crowd driving, intentions of pedestrians are partially observable: the vehicle can not directly observe pedestrians' navigation goals. 
One can only model pedestrians' intentions with a belief, which needs to be inferred from the action-observation history. Both planning and learning have to be performed in the enormous-scale belief space.

Our results in simulation show that \letsdrive outperforms either planning or imitation learning alone in both seen and novel environments. It successfully develops sophisticated driving skills, and enables the vehicle to achieve its long-term goal more safely and efficiently.




\section{Background and Related Work}

\subsection{Online POMDP Planning}
A major challenge in real-world planning tasks is \textit{partial observability}: system states are not known exactly and one can only receive observations from the world, which reveals some information about the true state. POMDP offers a principled way to handle partial observability and uncertainties in sensing and control.

Formally, a POMDP model is represented as a tuple $(S,A,Z,T,O,R)$, where $S$ represents the state space of the world, $A$ denotes the set of all possible actions, and $O$ represents the observation space. The transition function $T$ characterizes the dynamics of the world. When the robot takes an action $a$ at state $s$, the world transits to a new state $s'$ with a probability $T(s,a,s')=p(s'|s,a)$. After that, the robot receives an observation $z$ with probability $p(z|s',a)=O(s',a,z)$, and also a real-valued reward $R(s,a)$.

To perform planning, the robot maintains a belief $b$, represented as a probability distribution over $S$. POMDP planning searches for a policy $\pi:B\rightarrow A$ which prescribes an action $a$ that optimizes future values at each belief $b$. For infinite horizon POMDPs, the value of a policy $\pi$ at a belief $b$ is defined as the expected total discounted reward achieved by executing the policy $\pi$ from $b$ onwards:
\begin{equation}
 V_{\pi}(b)=E\left(\sum\limits_{t=0}^{\infty} \gamma^t R(s_t, \pi(b_t))~|b_0=b\right)
\end{equation}

Online POMDP planning interleaves planning and execution to efficiently handle complex tasks. 
Assume that the robot starts from an initial belief $b_0$. At each time step $t$, the planning computes an optimal action $a^*$ for the current belief, executes it immediately, and re-plans for the next time step. 
Online planning usually performs a look-ahead search from the current belief $b$. The planning constructs a belief tree and searches for an optimal policy $\pi^*$ that maximizes the value: $ V_{\pi^*}(b)=\max\limits_{\pi}\{V_{\pi}(b)\}$. 
The robot then executes the first action in the optimal policy and updates the current belief based on the action $a_t$ taken and the observation $z_t$ received. The belief update, denoted as $b_t=\tau(b_{t-1},a_t,z_t)$, is performed using the Baye's rule:
\begin{equation}
b_t(s')=\eta O(s',a_t,z_t)\sum\limits_{s\in S} T(s,a_t,s')b_{t-1}(s)
\end{equation}
where $\eta$ is the normalization constant. The new belief $b_t$ then becomes the entry of the next planning cycle.

POMDP planning suffers from the well-known ``curse of dimensionality'' and ``curse of history'' \cite{Kaelbling_1998}: the complexity of planning grows exponentially with the size of the state space and the planning horizon.
Two recent belief tree search algorithms, POMCP \cite{POMCP} and DESPOT \cite{DESPOT} made online POMDP planning practical for real-world tasks. Both of them use Monte Carlo simulations to sample transitions and observations. POMCP \cite{POMCP} performs Monte Carlo tree search (MCTS) on the belief space. The DESPOT algorithm \cite{DESPOT} samples a fixed set of scenarios and constructs a sparse belief tree conditioned under these scenarios. DESPOT further performs branch-and-bound pruning in the belief tree. Both POMCP and DESPOT are able to solve moderate-scale POMDP problems efficiently, while DESPOT achieves significantly better worst-case performance. DESPOT also offers a better chance for parallelization, enabling a massively parallel planner, HyP-DESPOT \cite{Hyp-despot}.



\subsection{Navigation or Driving in a Crowd}

\begin{figure*}
\centering
\includegraphics[width=0.6\textwidth]{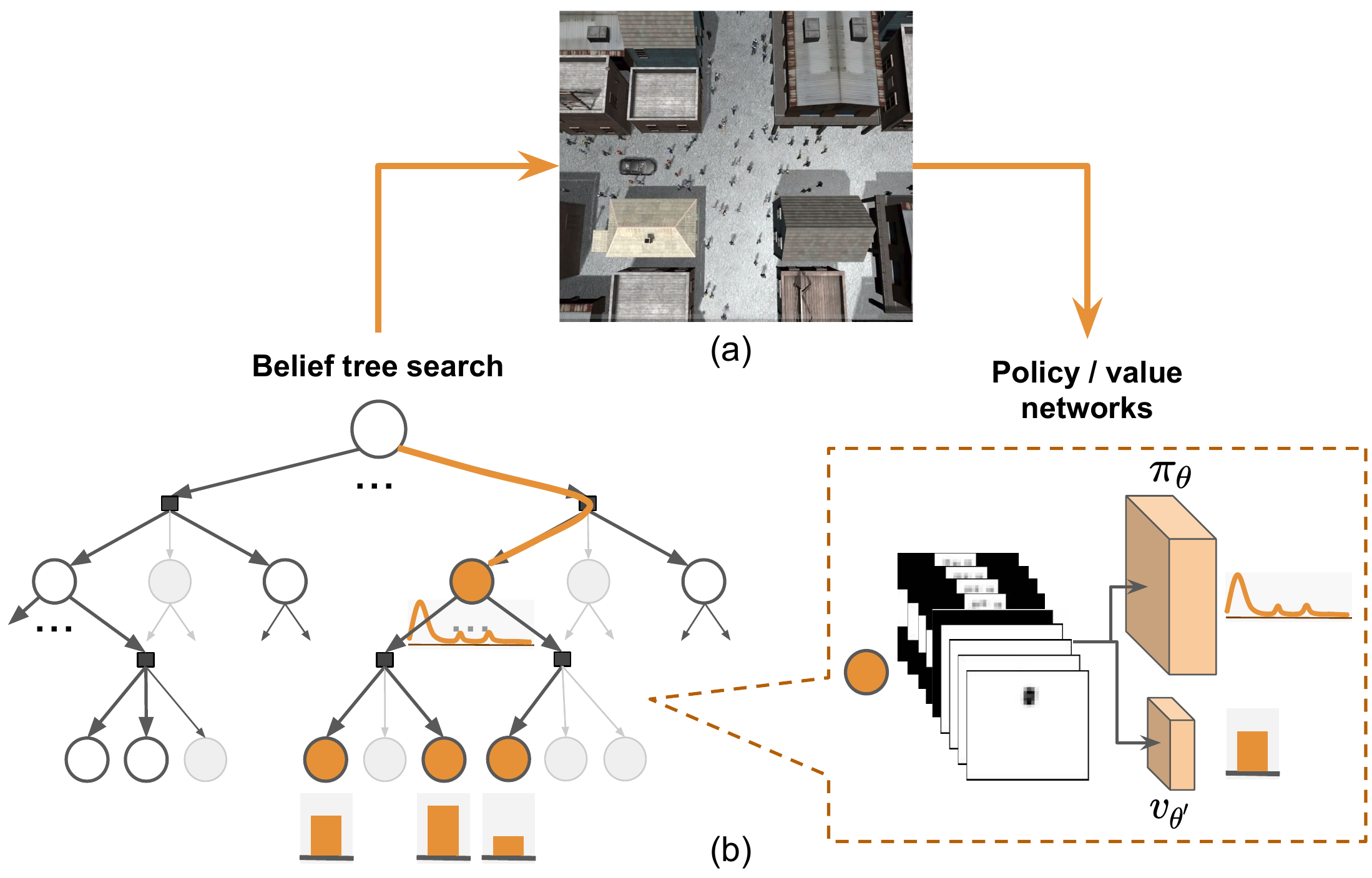}
\caption{
Overview of the two stages in \letsdrive. (a) ``Imitation'': training a policy network and a value network using a belief tree search expert, and (b) ``Improvement'': guide the belief tree search with the learned policy and value functions.
}
\label{fig:Overview}
\end{figure*}

Existing work in crowd driving or navigation fall into three main categories: local collision avoidance, learning-based, and planning-based approaches. 

One type of local collision avoidance algorithms is social force \cite{helbing1995social,lohner2010modeling,ferrer2013robot} that assumes pedestrians are driven by virtual attraction forces exerted by their navigation goals and repulsing forces exerted by obstacles. Social force-based algorithms are suitable for simulating crowds, but can easily be stuck in local optima when controlling robots. Another class are Velocity Obstacle (VO) \cite{fiorini1998motion} and RVO \cite{van2008reciprocal,van2011reciprocal,snape2011hybrid}-based methods. They compute command velocities by performing optimization in the feasible velocity space. A representative is ORCA \cite{ORCA}, that models homogeneous agents assuming that they perform collision avoidance with each other in a reciprocally optimal way. Recently, PORCA \cite{PORCA} improves the model to handle pedestrian-vehicle interactions.

Learning-based approaches fall into two categories. Some learn to predict pedestrian motions \cite{large2004avoiding, bennewitz2005learning} and use the predictions for navigation planning. These approaches lack a mechanism to handle uncertainties in the learned model and thus could be sensitive to prediction errors. Other methods directly learn a navigation policy using imitation learning \cite{long2017} or reinforcement learning \cite{chen2017a,chen2017b,fan2018}. These approaches enabled robots to successfully navigate among real pedestrians. 
However, learning from an ORCA expert limits the learned policy to imitate local collision avoidance behaviors. 
Instead, our expert (\secref{sec:il}) demonstrates global planning for the policy network.

Planning-based approaches suffer from high computation burden. To make the problem tractable, prior methods \cite{Bai_2015, PORCA} decoupled the driving actions and solved two sub-problems: a path planner to determine the steering angle, and a POMDP to control the acceleration. The decoupled approach significantly reduced the computational cost, but it also breaks the cooperation between steering and acceleration, which is often the key to sophisticated driving: human drivers often speed up and maneuver in the meantime to cut through others' way.

\subsection{Integrating Tree Search and Learning} 

\letsdrive is inspired by AlphaGO \cite{AlphaGO}, AlphaGO Zero \cite{AlphaGOZero}, and AlphaZero \cite{AlphaZero} that integrate Monte Carlo Tree Search with neural networks to learn board games like GO, Shogi, and chess from self-play. The idea has also been applied to Hexagon \cite{Hexgame}, another perfect information two-player game. Recently, DeepStack \cite{DeepStack} extends the idea to solve a partial information poker game using a value function based on the belief of the opponent's hands.

\letsdrive  faces two major  challenges in applying the scheme to real-world robotics tasks. First, instead of a single opponent, the vehicle interacts with many pedestrians in crowd driving. It is intractable to search over the actions of all pedestrians.
Instead, we use a motion model (\secref{sec:motion_model}) to predict pedestrians' motion, and condition the model on their intended navigation goals. 
Second, such pedestrian intentions are partially observable to the vehicle. We can only infer a belief over pedestrians' intentions from the interaction history. \letsdrive maintains the belief using Bayesian filtering \cite{sarkka2013bayesian}. Then, it plans in the belief space, and conditions the neural networks on history states.
Note that although DeepStack also includes a notion of belief, the immediate consequences of player's actions are deterministic. In crowd driving, transitions of pedestrians are highly uncertain due to their complex interactions, producing significantly more complex belief trees to search.

\section{\letsdrive}
\letsdrive integrates POMDP planning and deep learning to drive a vehicle among many moving pedestrians. 
The method (\figref{fig:Overview}) contains an \textit{imitation} stage and an \textit{improvement} stage. 

In the imitation stage (\figref{fig:Overview}(a)), \letsdrive uses neural networks to approximate the global planning behavior of a belief tree search expert. It generates demonstration trajectories in simulation to train a policy network and a value network, the architectures of which are designed using our domain knowledge in navigation. These networks take as input the current history state and the intention of the vehicle, and predict distributions on the vehicle's steering and acceleration, as well as the expected value of the policy. 

In the improvement stage (\figref{fig:Overview}(b)), \letsdrive uses the learned policy and value functions to guide the belief tree search, leveraging their prior knowledge to efficiently explore promising paths and avoid searching deep subtrees.
The guided search constructs an improved policy over the learned one, and corrects critical errors made by the neural networks.
At runtime, \letsdrive executes the guided search to drive a vehicle.

In this section, we present our approach in detail, including the POMDP model, the architectures and the learning procedure of the neural networks, as well as how they are used to guide the belief tree search.

\subsection{The POMDP model} \label{sec:model}
\subsubsection{State and Observation Modeling}
A state in crowd driving consists of the vehicle state and the states of $20$ nearest pedestrians.
The vehicle state consists of its 2D position, heading direction, and instantaneous speed. 
A pedestrian state contains its position and speed,
as well as the intention represented as his/her navigation goal. 
We assume that positions and velocities are fully observable, and pedestrians adopt a finite set of possible navigation goals, \ie, destination locations, as known for each environment.
However, their actual intentions are partially observable to the vehicle, and thus need to be modeled as beliefs and must be inferred from the interaction history.

\subsubsection{Action Modeling} \label{subsubsec:action-modeling}
Sophisticated driving relies on collaborations of steering and acceleration. Therefore, we plan in a two-dimensional joint action space comprising the steering of the front wheel, discretized using a resolution of 5 degrees, and the acceleration of the vehicle, containing three discrete values, \accel, \decel, and \maintain. 

\subsubsection{Transition Modeling} \label{sec:motion_model}
The transition function models the dynamics of the vehicle and nearby pedestrians. We use a bicycle model to transit the vehicle according to specific steering and acceleration commands. For pedestrians, we use PORCA \cite{PORCA} to predict their motion conditioned on the inferred intentions. PORCA assumes that pedestrians optimize their instantaneous velocities to approach the intended goals and comply with collision avoidance constraints in the meantime. The model further assumes that interactions among pedestrians are ``reciprocal'', and vehicles and pedestrians share different responsibilities in reciprocal collision avoidance due to their own limitations.

\subsubsection{Reward Modeling}
The reward function encourages the vehicle to drive safely, efficiently, and smoothly. For safety, we give a large penalty $R_\mathrm{col}=-1000\times(\speed^2 + 0.5)$, varying with the driving speed \speed, to the vehicle if it collides with any pedestrian. For efficiency, we assign a small cost $R_\mathrm{time} = -0.1$ to each time step and issues a reward $R_\mathrm{goal} = 0$ to the vehicle when it reaches the goal. For smoothness of the drive, we add a small penalty $R_\mathrm{acc} = -0.1$ if the action performs \accel~or \decel, to penalize the excessive speed changes.

\subsection{Learning from Tree Search}\label{sec:il}

In the imitation stage, we train our neural networks using a belief tree search expert. Note that it is intractable to directly solve the POMDP in \secref{sec:model}. Instead, we use \pomdpbaseline \cite{Bai_2015,PORCA} as the expert.
The algorithm decouples the planning of steering and acceleration to achieve real-time performance. At each time step, it plans a path using Hybrid A* \cite{stanley2006robot} and restricts POMDP to control only the accelerations along the path. 

\subsubsection{Dataset} 
We first use \pomdpbaseline to drive the vehicle in simulation and generate demonstration trajectories. The trajectories are sparsely sampled to form a data set $D$. Each data point in $D$, denoted as $(H, I, \alpha, a, V)$, where the history state $H$ and the reference path $I$ are inputs to the neural networks, and the actions $\alpha$, $a$ and the value $V$ are labels to predict. Particularly, $H = (H_c, H_e)$ represents a fixed-length history of the vehicle and the environment (moving pedestrians and static obstacles), respectively. This history not only encodes the current state and the dynamics of the involved agents, but also reveals the intentions of pedestrians. The reference path $I$, generated by Hybrid A* considering only static obstacles, represents the intention of the robot vehicle.
Finally, labels in the data point record the steering $\alpha$ and the acceleration $a$ chosen by the expert, and the accumulative reward $V$ that the vehicle collected from the current time step till the end of the trajectory.




\subsubsection{Training}
Using this dataset, \letsdrive trains a policy network parameterized by $\theta$:
$$\pi_\theta: ~ (H_c, H_e, I) \rightarrow \left( f(\alpha), f(a) \right) $$
where $f(\alpha)$ and $f(a)$ are distributions over the steering and the acceleration of the vehicle. We also fit a value network parameterized by $\theta'$:
$$v_{\theta'}: ~ (H_c, H_e, I) \rightarrow V $$
where $V$ is the predicted value of the history state.


The policy network $\pi_\theta$ and the value network $v_{\theta'}$ are trained separately using supervised learning. The loss functions, $l(\theta, D)$ and $l(\theta', D)$, measure the errors in action and value predictions, respectively:
\begin{eqnarray} \label{equation::action_loss}
l(\theta, D) &=& H_\alpha(\theta, D) + H_a(\theta, D)\\
l(\theta', D) &= &MSE_V(\theta', D)
\end{eqnarray}
where
\begin{eqnarray} \label{equation::action_loss1}
H_\alpha(\theta, D) &=& - \frac{1}{N}\sum_{i}^N \log \pi _{\theta}(\alpha_i | H_i, I_i)\\
H_a(\theta, D) &=& - \frac{1}{N}\sum_{i}^N \log\pi_{\theta}(a_i | H_i, I_i) \\
MSE_V(\theta', D) &=& \frac{1}{N}\sum_{i}^N(v_{\theta'}(H_i, I_i)-V)^2
\end{eqnarray}
Here, $N$ is the size of the data set, $H$ represents the cross-entropy loss \cite{CrossEntropy} of action predictions, and MSE denotes the mean square error of value predictions. 
All inputs $(H_c, H_e, I)$ are initially encoded as $1024\times1024$ images and down-sampled to $32\times32$ using Gaussian pyramids \cite{adelson1984pyramid} before inputting to the neural networks.

\subsection{Guided Belief Tree Search} \label{sec:search}
In the improvement stage, \letsdrive uses HyP-DESPOT \cite{Hyp-despot}, a massively parallel belief tree search algorithm, to plan vehicle motions. \letsdrive incorporates the prior knowledge learned in the policy and value networks into the heuristics of HyP-DESPOT in order to search efficiently within the limited planning time.

For completeness, we provide a brief summary of HyP-DESPOT.
See~\cite{Hyp-despot} for details.
HyP-DESPOT samples a small set of \nscen scenarios as representatives of
the stochastic future.
Each scenario, $\scenario=(\sinit, \randnum_1,\randnum_2,...)$, contains a
sampled initial state \sinit and random numbers $\randnum_1,\randnum_2,...$ that determinize the outcomes of future actions and observations.
HyP-DESPOT iteratively constructs a sparse belief tree conditioned on the sampled scenarios. 
Each node \node of the tree contains a set of scenarios \scenariosetnode{\node},
whose starting states represents a belief. The tree starts from an initial belief. It branches on all actions, but
only on observations encountered under the sampled scenarios.  
In each trial, HyP-DESPOT starts from the root node \rootnode and searches a single path down to
expand the tree. At each node along the path, HyP-DESPOT chooses an action branch
and an observation branch according to the heuristics computed using an upper bound and a lower bound value, \ubsymbol and \lbsymbol. It ends a trial when it is no longer beneficial, and immediately backs-up the upper bounds and lower bounds to update the root. The search terminates until the gap between the upper and
lower bounds at the root is sufficiently small or the planning time limit is reached.

\letsdrive integrates HyP-DESPOT with the neural networks to solve the joint-action POMDP efficiently.
During the forward search, \letsdrive uses the learned policy to bias action selections, so that the search prefers to explore actions chosen by the expert:
\begin{equation}
a^*=\argmax_{a\in A} \left \{ u(b,a) + c \pi_{\theta}(a | x_b)\sqrt{\frac{N(b)}{N(b,a)+1}} \right \}  
\label{Eqn:select_ub}
\end{equation}
Here, the first term exploits actions with higher upper bound values \uba{\node}{\act}. The second term is the exploration bonus considering the visitation counts of the node, \nvisit{\node}, and that of the action \act under the node, \nvisita{\node}{\act}. It also depends on the prior probability $\pi_{\theta}(a | x_b)$, of choosing action $a$ at the history state $x_b$, as suggested by the policy network $\pi_{\theta}$. 
In effect, the heuristics trade off the exploitation of high-quality branches and the exploration of less-visited ones with high prior probabilities.
The constant scaling factor $c$ controls the desired level of exploration and can be set empirically.

When encountering a new belief node, \letsdrive queries the value network to provide a tight estimation of the lower bound and expedite the convergence of the search:
\begin{equation}
l_0(b)=v_{\theta'}(x_b)
\end{equation}
where $v_{\theta'}(x_b)$ is the value predicted by the value network $v_{\theta'}$ at history state $x_b$. This tight estimation can otherwise only be acquired by sufficiently searching the corresponding subtree.

\figref{fig:Overview}(b) illustrates the guided belief tree search in \letsdrive.


\begin{figure}
\centering
\includegraphics[width=0.47\textwidth]{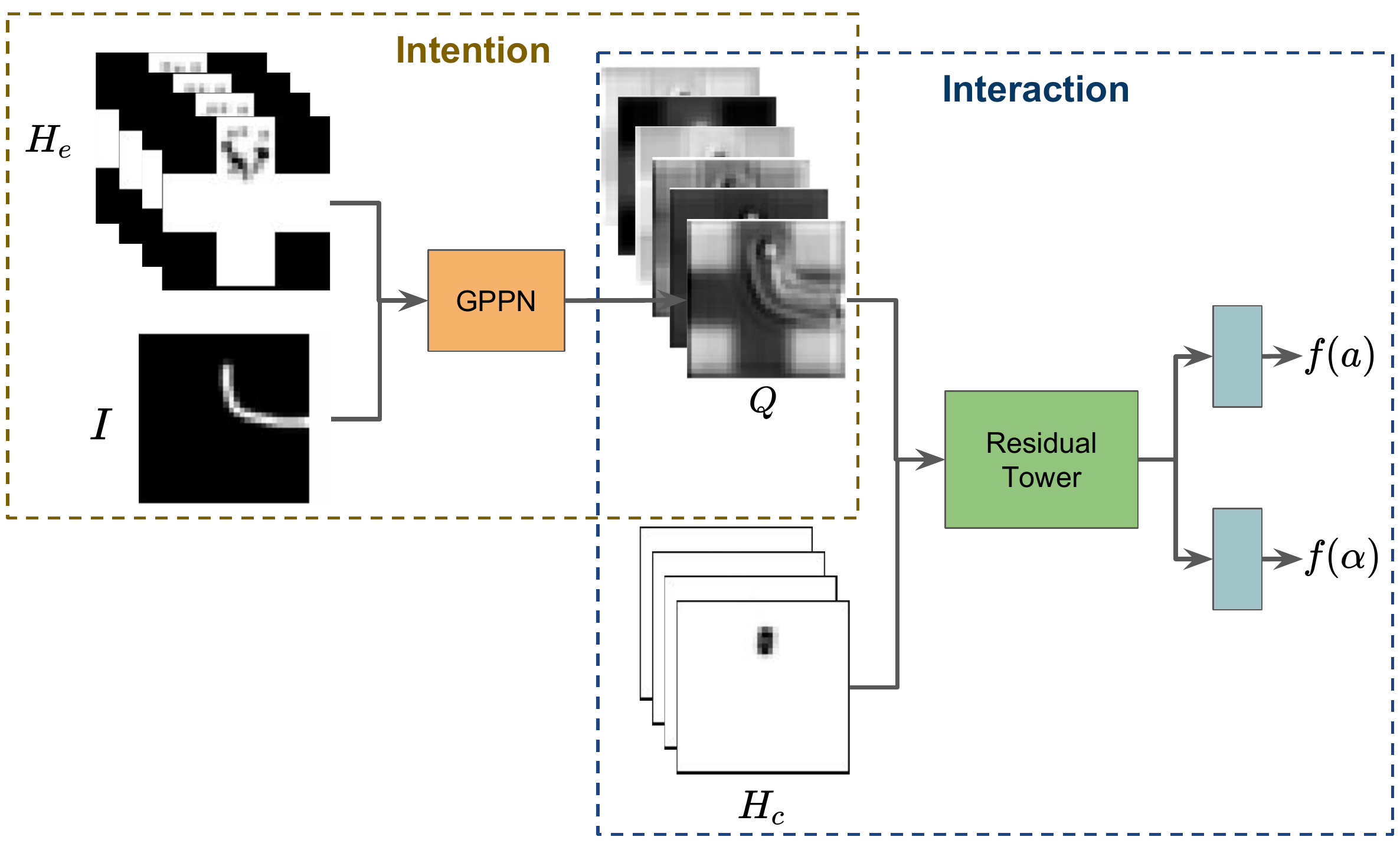}
\caption{
The architecture of the \letsdrive policy network with an intention module and an interaction module. The intention module takes as input the dynamic environment $H_e$ and the intended path $I$ of the vehicle, and outputs action-values $Q$ for all locations on the map. Then, the interaction module takes these action-values, and outputs action probabilities according to the current location and the history of the vehicle.
}
\label{fig:Network-architecture}
\end{figure}

\subsection{Neural Network Architecture} \label{sec:architecture}

\letsdrive requires the neural networks to represent high-quality policies with low computational complexity.
We design the neural network architectures according to the domain knowledge in planning. An optimal policy should convey two key aspects: conduct the vehicle's intention and interact with pedestrians to avoid collisions. Therefore, our policy network first computes an initial plan using the vehicle's intention, then refines the policy to incorporate interactions with another neural network module.

Particularly, our policy network has an ``intention'' module and an ``interaction'' module. The intention module takes as input the dynamic environment and the intention of the vehicle, and outputs \textit{action-values}, \ie, the value to be achieved if the vehicle takes a specific action at the current step, for all locations on the map. Then, the interaction module takes these action-values, and outputs action probabilities according to the current location and the history of the vehicle. 

The architecture of the intention module follows the spirit of Value Iteration Networks (VINs) \cite{VIN}, which embeds an MDP model \cite{MDP} and the value iteration algorithm \cite{VI} in a recurrent neural network to perform end-to-end training for the model and the algorithm simultaneously.
Formally, we use a Gated Path Planning Network (GPPN) \cite{GPPN} which is an improved variant of VIN that provides faster training and better generalization. 
It first maps the history of the environment $H_e$ and the vehicle's intended path $I$ into reward images, then performs recurrent convolutions using an LSTM \cite{LSTM} on the reward images to perform general value iterations on the map.
The intention module outputs a set of action-value images $Q$. Channels of $Q$ correspond to actions encoded in a learned latent space, and each pixel in $Q$ corresponds to a location on the map.

The interaction module uses these $Q$ images to determine the vehicle's action according to the vehicle's current history. 
A typical way is to use a simple attention mechanism for this: extract the $Q$ values at the vehicles current position and calculate the best actions from the extracted action-values \cite{VIN}. However, this approach fails in dynamic environments, because values on the map will change through time.
Instead, \letsdrive uses a Convolutional Neural Network (CNN) \cite{CNN} to account for the complex interactions between the vehicle and the dynamic environment. The interaction module first stacks the $Q$ images with the current history of the vehicle $H_c$ and pass them through a residual tower concatenating $11$ convolutional layers. 
The images are first processed with $2$ residual blocks \cite{ResNet} with $32$ filters with kernel size $3\times3$ and stride $1$. 
The processed images are then down-sampled to $16\times16$ using a convolutional layer with $64$ filters with kernel size $7\times7$ and stride $2$ followed by batch normalization and rectifier nonlinearity. 
The down-sampled images further pass through $3$ residual blocks with $64$ filters with kernel size $3\times3$ and stride $1$ and finally input to two action heads to produce steering and acceleration commands. 
The heads first apply a 50\% dropout on the images, Then, it uses $4$ filters of kernel size $1\times1$ to obtain 4 feature images, flattens them as a 1-dimensional vector and uses a fully-connected layer to produce the final predictions. 
\figref{fig:Network-architecture} illustrates the high-level architecture of this policy network.

The value network in \letsdrive needs to be queried much more frequently in the belief tree search (\secref{sec:search}) than the policy network, thus needs to be computationally light-weighted. Therefore, we removed the GPPN module from the value network. Instead, it directly stacks the dynamic environment $H_e$, the history of the vehicle $H_c$ and the intended path $I$ and processes them with a similar but smaller residual tower as that in the interaction module. This residual tower only has $2$ residual blocks after the down-sampling layer and uses only 16 filters in the last $5$ convolutional layers. The output images from the residual tower are passed through a value head, which is similar to the action heads in structure, to produce value predictions for the current history state.

\begin{figure}
\centering
\includegraphics[width=\columnwidth]{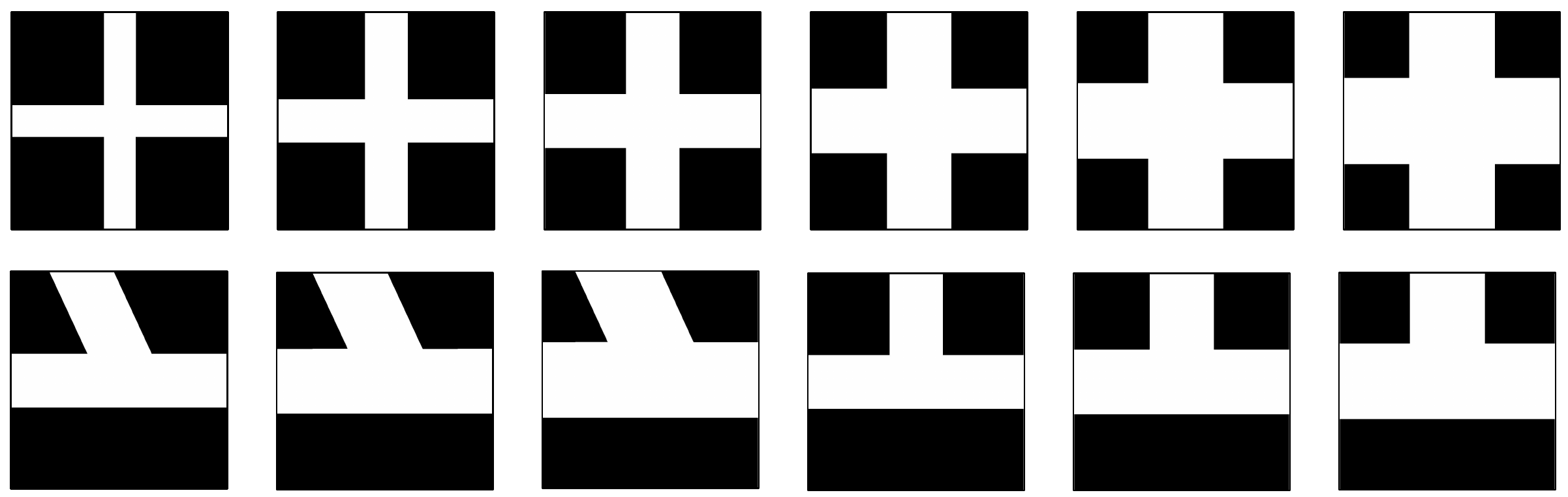}
\caption{
Driving maps with varying road widths and layouts used for training: black regions represent static obstacles (\ie, buildings) and white regions represent roads that the vehicle and pedestrians can drive or walk on. Road widths vary from 8 meters to 16 meters.
}
\label{fig:Maps}
\vspace*{-0.5cm}
\end{figure}

\begin{figure}
\centering
\begin{tabular}{ccc}
\fbox{\includegraphics[width=0.2\columnwidth]{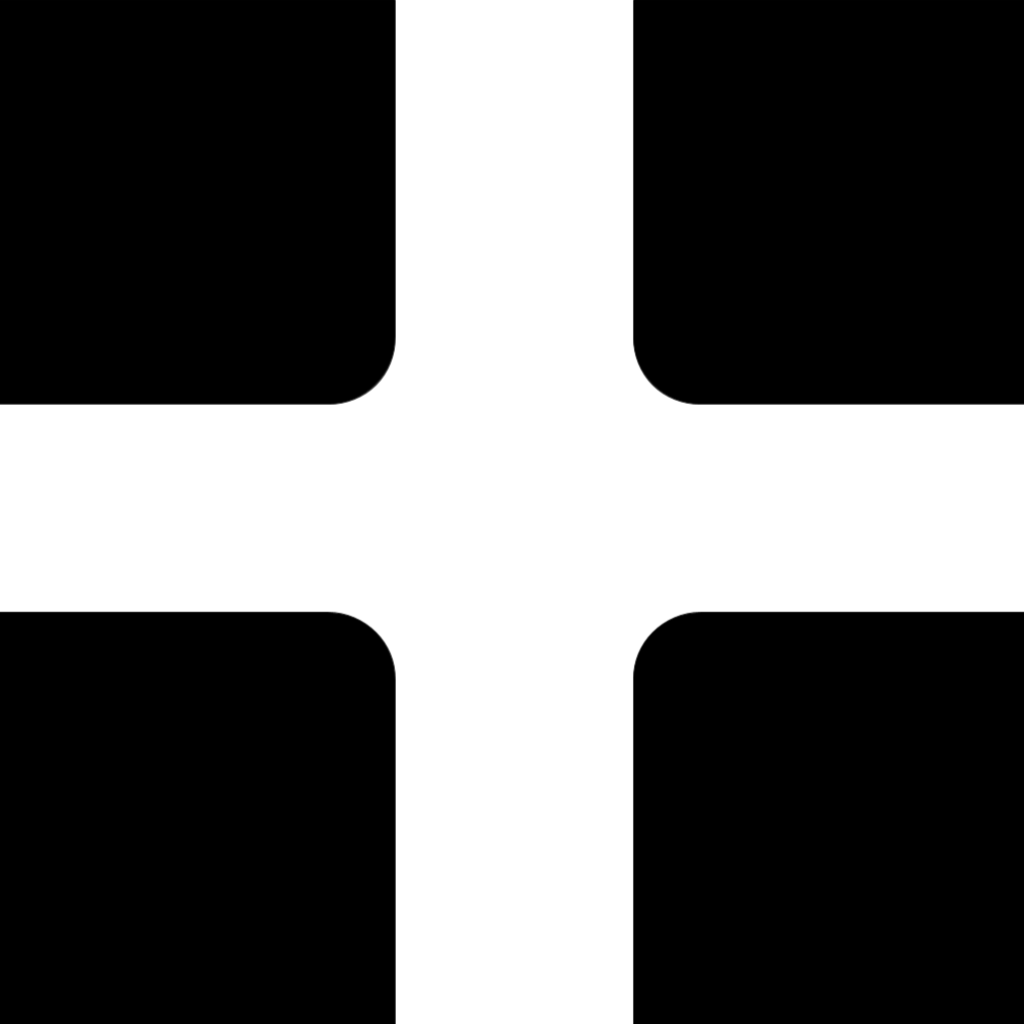}} &
\fbox{\includegraphics[width=0.2\columnwidth]{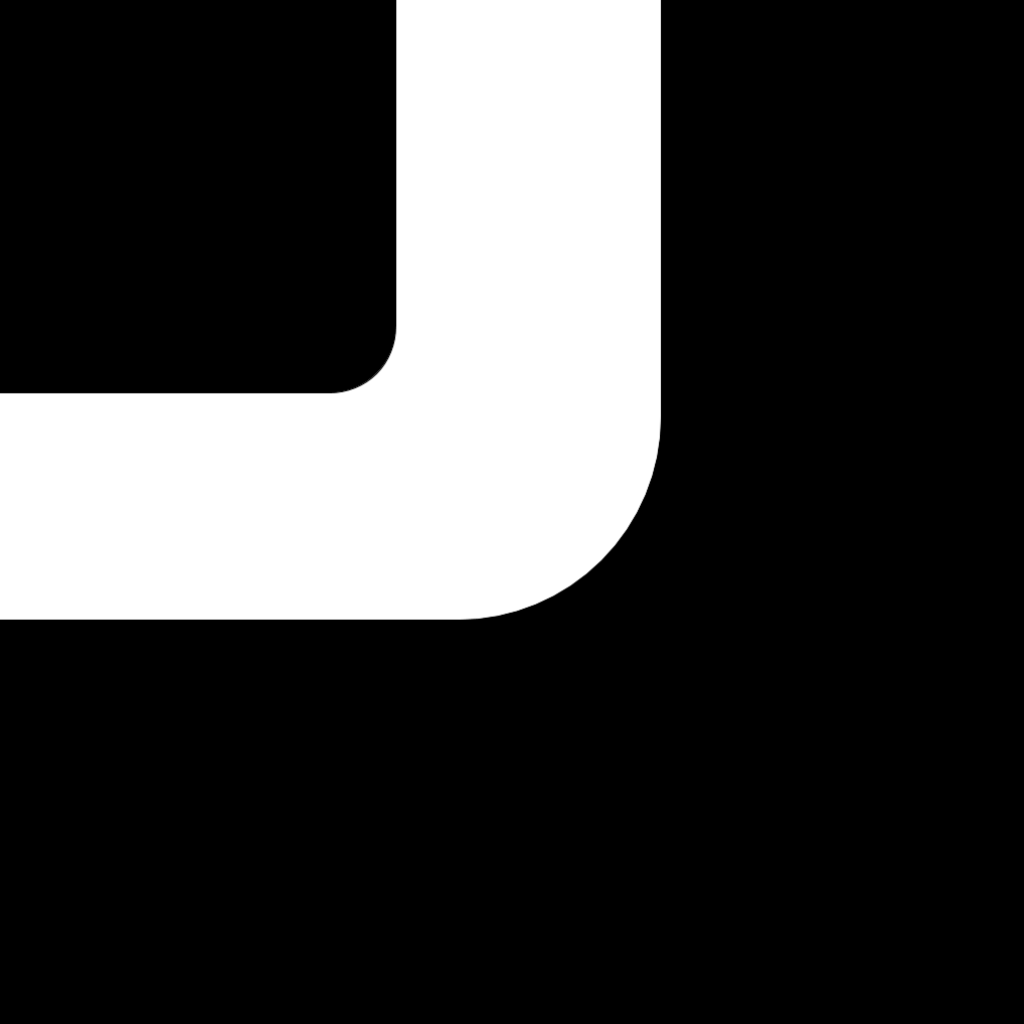}}&
\fbox{\includegraphics[width=0.2\columnwidth]{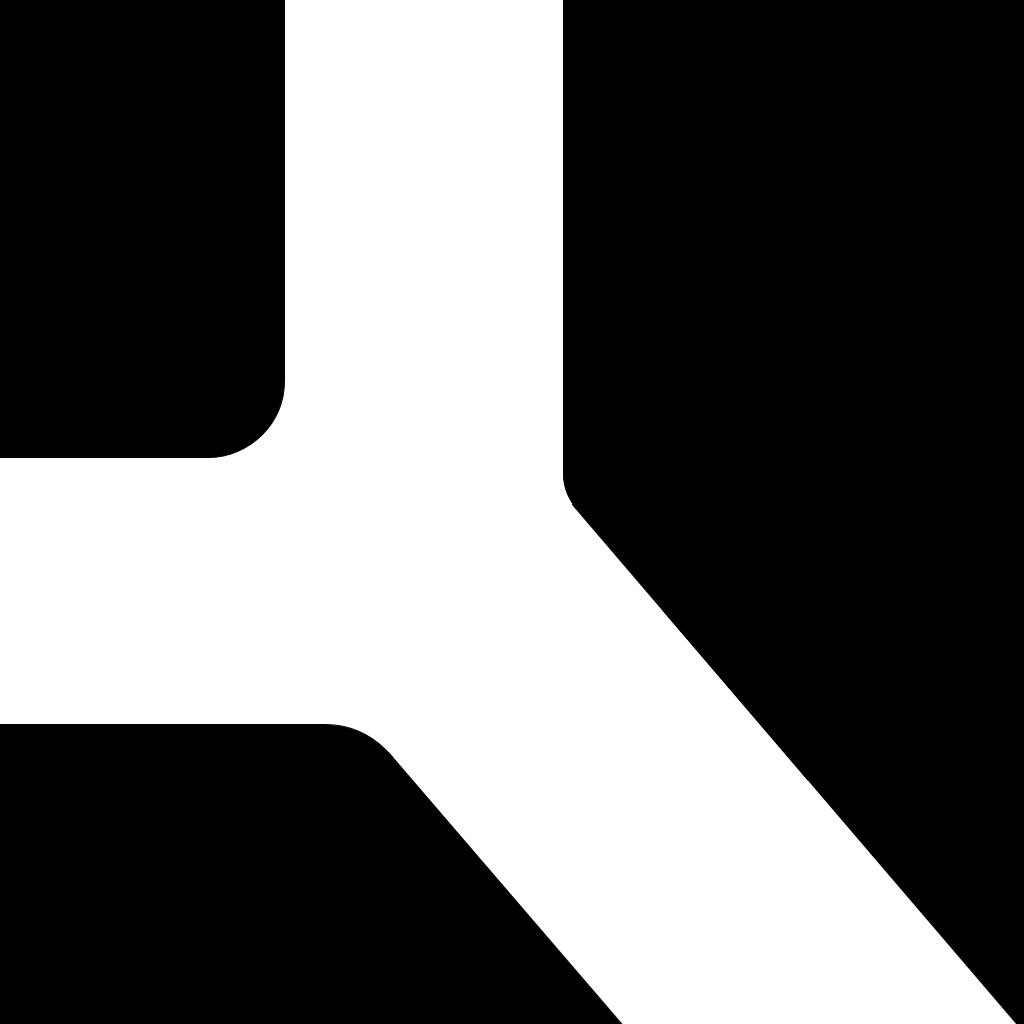}} \\
(\subfig{1}) & (\subfig{2}) & (\subfig{3}) \\
\end{tabular}
\caption{
Maps for generalization test of \letsdrive.
}
\label{fig:testmaps}
\vspace*{-0.5cm}
\end{figure}

\section{Results and Discussions}
We used \letsdrive to drive a vehicle in simulated crowds to analyze its performance. Our results show that global planning is important for driving in a crowd, and our policy network can efficiently learn the behavior of global planning. By further integrating the learned policy with the belief tree search, \letsdrive achieves superior driving performance in simulation, compared to local collision avoidance, POMDP planning, and imitation learning.

\subsection{Driving Simulation}

The driving simulator is built using a sophisticated game engine Unity3D. We built 12 different driving maps of size $40 m\times 40 m$ to evaluate \letsdrive (\figref{fig:Maps}). Six of them are crossroad maps with different road widths, and the other six are junctions. Roads in the maps are occupied by crowds of people, each pedestrian has his/her own destinations unknown to the robot vehicle a~priori. To generate realistic scenes, we simulate the crowds using PORCA, which has been shown to produce accurate predictions of pedestrian motion in the presence of vehicles \cite{PORCA}.

In all driving scenarios, we randomized the initial positions and navigation goals of pedestrians, as well as the starting positions and initial directions of the vehicle. In effect, the vehicle faces a new environment in each drive.

\subsection{Performance comparisons} \label{sec:comparison}

\begin{table*}
\centering
  \caption{Comparisons on the driving performances of \porcabaseline, \hypbaseline, \pomdpbaseline, \ilbaseline, and \letsdrive in training maps.}
  \begin{tabular}{ l c c c c}
    \toprule
    & Collision rate & Success rate & Time-to-goal & \# Decelerations\\ \hline
    \porcabaseline& 0.171 & 0.573 & 37.2 $\pm$ 0.67 & 24.6 $\pm$ 1.16 \\ \hline
    \hypbaseline & 0.016 & 0.774 & 33.5 $\pm$ 0.51 & 53.8 $\pm$ 2.04 \\ \hline
    \pomdpbaseline & 0.005 & 0.998 & 43.2 $\pm$ 0.73 & 38.8 $\pm$ 1.14 \\ \hline
    \ilbaseline & 0.012 & 0.946 & 43.5 $\pm$ 0.96 & 46.4 $\pm$ 0.89 \\ \hline
    \letsdrive & \textbf{0.002} & \textbf{0.998} & \textbf{29.6 $\pm$ 0.41} & \textbf{18.2 $\pm$ 0.54} \\
    \bottomrule
  \end{tabular}
   \label{tab:trainmap_results}
\end{table*}


\begin{table*}[]
\centering
\caption{Comparisons on the driving performances of \pomdpbaseline, \ilbaseline, and \letsdrive in the three test maps.}
\begin{tabular}{lccccc}
\toprule
Test Map              		& Algorithm                      & Collision rate                  & Success rate                  & Time-to-goal                    & \# Decelerations \\
\hline
\multirow{3}{*}{Map 1}		
                      		&  \pomdpbaseline 				 & 0.003                           &  1.0 				        & 44.5 $\pm$ 0.81             & 41.6 $\pm$ 1.15 \\
                          &  \ilbaseline             & 0.003                           & 0.99                    & 38.2 $\pm$ 0.78                & 26.5 $\pm$ 0.85 \\
                          &  \letsdrive              &  \textbf{0.001}           & \textbf{1.0}                         &  \textbf{28.2 $\pm$ 0.47}                  & \textbf{15.7 $\pm$ 0.46} \\
\hline
\multirow{3}{*}{Map 2}      
                            &  \pomdpbaseline        &  0.017             & 0.98                      & 52.8 $\pm$ 1.10                  & 56.2 $\pm$ 1.57 \\
                            &  \ilbaseline           &  0.005    &  0.96           & 44.8 $\pm$ 1.37                   & 32.8 $\pm$ 0.98 \\
                            &  \letsdrive            & \textbf{0.005}                           & \textbf{1.0}                 &  \textbf{28.9 $\pm$ 0.65}                           & \textbf{16.6 $\pm$ 0.51} \\
\hline
\multirow{3}{*}{Map 3}      
                            &  \pomdpbaseline 				 & 0.002                   &  \textbf{1.0} 				   & 36.5 $\pm$ 0.70                 & 32.2 $\pm$ 1.07 \\
                            &  \ilbaseline             &  \textbf{0.0}             & 0.94                          & 31.6 $\pm$ 0.64                & 23.6 $\pm$ 0.86 \\
                            &  \letsdrive              & 0.0007                            & 0.99                         & \textbf{24.6 $\pm$ 0.50}                          & \textbf{11.2 $\pm$ 0.40}\\
\bottomrule                        
\end{tabular}
    \label{tab:testmap_results}
\end{table*}

We tested the driving performance of five algorithms, including \Porcabaseline, \Hypbaseline, \Pomdpbaseline, \Ilbaseline, and \letsdrive. 
\porcabaseline performs local collision avoidance for the vehicle. Similar to PORCA \cite{PORCA} for pedestrians, \porcabaseline directly controls the velocity of the vehicle. To generate non-holonomic vehicle motions, it constrains the command velocities around the vehicle's current heading direction, and uses a pure-pursuit controller to execute the velocity. 
\hypbaseline applies HyP-DESPOT to directly solve the POMDP in \secref{sec:model}. To generate reasonable behaviors within the limited search time, we augmented the reward function to encourage the vehicle drive near the global path. \pomdpbaseline is the training expert that decouples the planning for steering and acceleration. It also uses HyP-DESPOT to plan for accelerations.
\ilbaseline executes our policy network that imitates the \pomdpbaseline expert. The proposed algorithm, \letsdrive, executes the belief tree search guided by the neural networks. In the following experiments, \letsdrive, \pomdpbaseline, and \hypbaseline use $0.3s$ of planning time (3 Hz frequency), while \ilbaseline and \porcabaseline generate control commands at $10$ Hz. 

The driving performance of the algorithms is measured using safety, indicated by the portion of trials where the vehicle collides with pedestrians or static obstacles (\textit{collision rate}), and efficiency, measured by the rate of reaching the goal within 2 minutes (\textit{success rate}) and the average time used to reach the goal in successful trials (\textit{time-to-goal}). We further include the number of decelerations (\textit{\# decelerations}) to measure the smoothness of a drive. \tabref{tab:trainmap_results} shows the average performance of around 1000 drives in simulation for each tested algorithm.

We first conclude that local collision avoidance is insufficient for driving in a crowd. \porcabaseline suffers in both safety and efficiency of driving as compared to planning, imitation learning, and \letsdrive. It often fails to reach the goal and frequently collides with pedestrians. This is due to the incapability of multi-step look-ahead, which is crucial for driving in a dynamic environment. 

\hypbaseline drives much more safely than \porcabaseline because it simulates the dynamics of the vehicle and pedestrians using forward search. However, due to the complexity of searching in the joint action space, the algorithm often fails to generate long-term plans.
Consequently, the vehicle misses the goal in around 22 \% of drives and generates jerky motions with frequent decelerations. In contrast, \pomdpbaseline achieves much better real-time performance.

The performance of \ilbaseline, matching up with its expert \pomdpbaseline, shows that our policy network (\secref{sec:architecture}) learns effectively from global planning. 

Finally, by integrating belief tree search with the learned policy and value functions, \letsdrive significantly outperforms \pomdpbaseline and \ilbaseline, achieving the lowest collision rate, and reducing the traveling time by roughly $30$ \% and the number of decelerations by more than $50$ \%.







\subsection{Generalization}
Since \letsdrive uses neural networks in belief tree search, it is important that the guided search generalizes to unseen environments. 
Results in \secref{sec:comparison} show that \letsdrive generalizes very well on randomized positions and intentions of pedestrians. 
In this section, we further built 3 novel maps (\figref{fig:testmaps}) to inspect the generalization capability of \letsdrive. The test maps have different layouts of static obstacles, which also affect the distribution and dynamics of pedestrians. 
\tabref{tab:testmap_results} shows the driving performance of algorithms on the test maps. 

The first test map has a similar crossroad layout as training maps (0)-(6) but has novel road positions, widths, and corner shapes. Both \letsdrive and \ilbaseline maintain their high performance as in the training maps. Among them, \letsdrive with guided belief tree search achieves the best safety, efficiency, and smoothness. 

The second and third test maps have very different road layouts from the training maps. \letsdrive still achieves significantly higher performance than either planning or imitation learning alone. Noticeably, the success rate of imitation learning decreases in the third map because the layout is significantly different. In contrast, \letsdrive recovers the performance by applying additional search upon the learned networks, achieving $99\%$ of success rate, and improving the efficiency by $22\%$ and the smoothness by $53\%$, approximately.










\subsection{Case Studies}

\begin{figure*}
\centering
\begin{tabular}{ccccc}
\Fbox{\includegraphics[width=\fw]{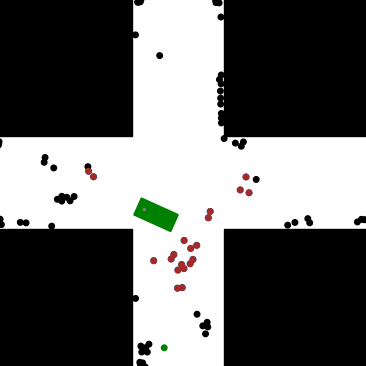}} &
\Fbox{\includegraphics[width=\fw]{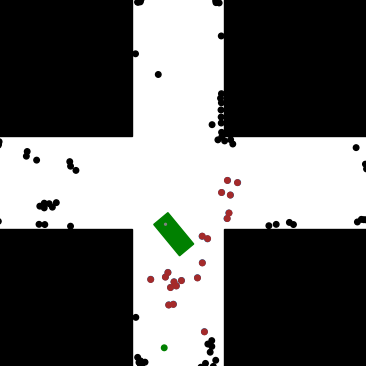}} &
\Fbox{\includegraphics[width=\fw]{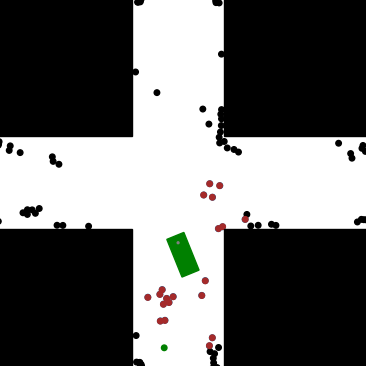}} &
\Fbox{\includegraphics[width=\fw]{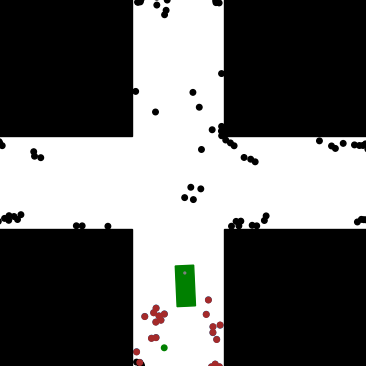}} &
\Fbox{\includegraphics[width=\fw]{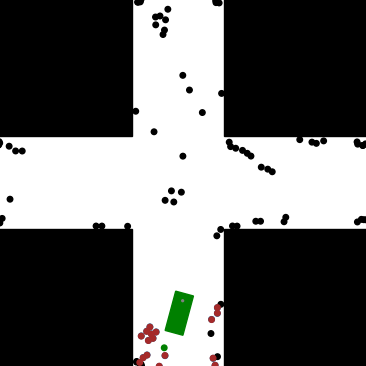}} \\
(\subfig{a}) & (\subfig{b}) & (\subfig{c}) & (\subfig{d}) & (\subfig{e}) \\
\end{tabular}
\caption{
\letsdrive case study: The vehicle moving south maneuvers to by pass pedestrians. The vehicle and its destination are shown in green. Pedestrians considered in the belief tree search are shown in brown color.
}
\label{fig:cases1}
\end{figure*}


\begin{figure*}
\centering
\begin{tabular}{cccc}
\Fbox{\includegraphics[width=\fw]{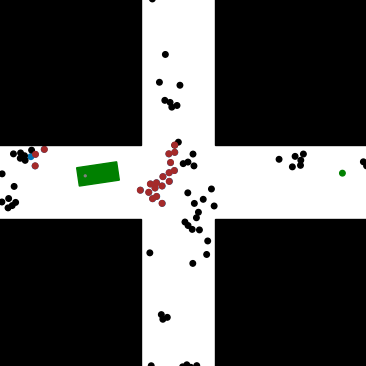}} &
\Fbox{\includegraphics[width=\fw]{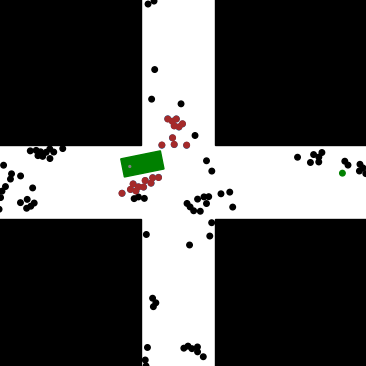}} &
\Fbox{\includegraphics[width=\fw]{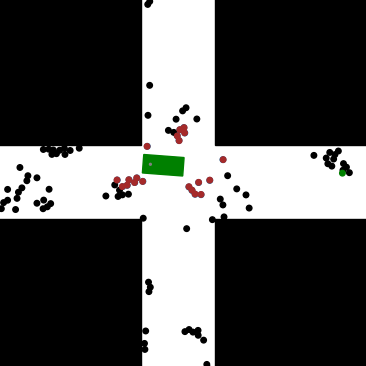}} &
\Fbox{\includegraphics[width=\fw]{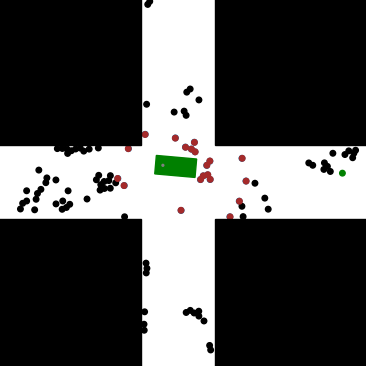}} \\
(\subfig{a}) & (\subfig{b}) & (\subfig{c}) & (\subfig{d}) \\
\Fbox{\includegraphics[width=\fw]{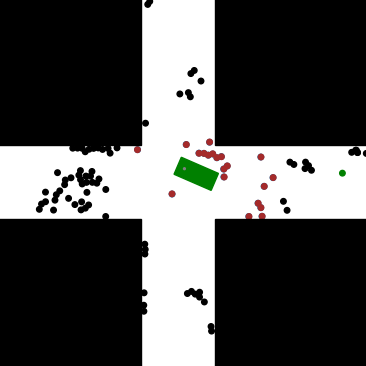}} &
\Fbox{\includegraphics[width=\fw]{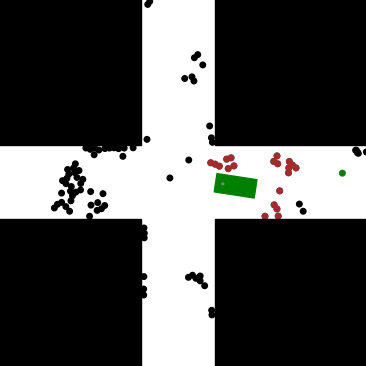}} &
\Fbox{\includegraphics[width=\fw]{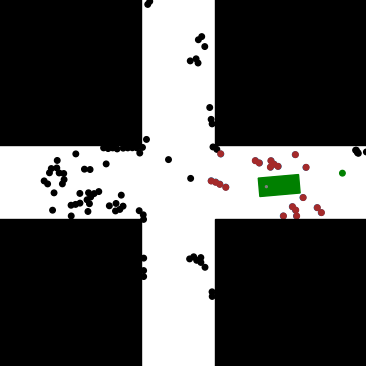}} &
\Fbox{\includegraphics[width=\fw]{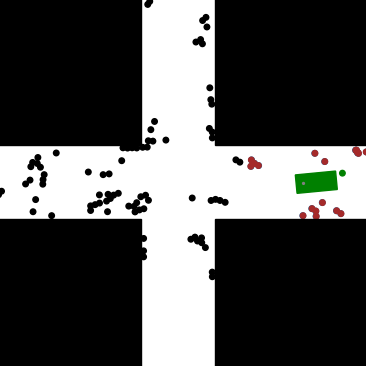}} \\
(\subfig{e}) & (\subfig{f}) & (\subfig{g}) & (\subfig{h}) \\
\end{tabular}
\caption{
\letsdrive case study: the vehicle (heading east) makes its way through the crowd by interacting with pedestrians. The vehicle and its destination are shown in green. Pedestrians considered in the belief tree search are shown in brown color.
}
\label{fig:cases3}
\end{figure*}

Human drivers reveal sophisticated driving behaviors in daily life: they drive around others who block their way, and can interact with pedestrians to make their way through even in very limited space. These behaviors, however, are hard to acquire for autonomous vehicles. In this section, we demonstrate that \letsdrive developed sophisticated driving behaviors after performing sufficient search.

In \figref{fig:cases1}, a group of pedestrians walk in right-front of the vehicle and blocks the way. Understanding that they intend at a similar destination as the vehicle, \letsdrive steered the vehicle around the crowd to take a longer, but more efficient path.

In \figref{fig:cases3}, many pedestrians walk near the vehicle and block its way and the free-space has only narrow passages. \letsdrive can still drive efficiently across the crowd by interacting with pedestrians with local maneuvering. 

Check this video \href{https://youtu.be/oghGK3QJFVo}{https://youtu.be/oghGK3QJFVo} for real-time driving cases in the Unity simulator.

\section{Conclusion}

This work addresses the challenge of autonomous driving among many                                   
pedestrians. \letsdrive leverages the robustness of planning and the runtime                        
efficiency of learning. It integrates them to advance the performance of                            
both. \letsdrive applies a two-stage approach. It learns a policy by imitating                      
the belief tree search and then guides the belief tree search using the                             
learned policy and its value function. \letsdrive outperforms the planning or the                         
learning approach alone in both safety and efficiency, and                                
demonstrates sophisticated driving behaviors in simulation. 

There are two main limitations in our current work. First,                             
the efficiency of online search is constrained by the complexity of the                         
learned  neural networks. To further improve performance, we plan to                            
investigate more efficient policy and value function representations.                    
Second, it remains  a  challenge to generalize the learned policies
and value functions over diverse maps and
agent behaviors.  We plan to expand the training environment with rich
real-world maps and more sophisticated agent motion models.

\section*{Acknowledgments}                                                                         
We want to thank Huaiyu Liu for help in implementing the neural networks for tree search.
This work is supported in part by the MoE AcRF grant 2016-T2-2-068 and National                            
Research Foundation Singapore through the SMART IRG program.
 \clearpage
\newpage



\bibliographystyle{ieeetr}      
\bibliography{RSS2019Cai}

\end{document}